# Design of an Intelligent Vision Algorithm for Recognition and Classification of Apples in an Orchard Scene

Hamid Majidi-Balanji [*], Alaeedin Rahmani Didar and Mohamadali Hadad Derafshi [**]

[*] Graduate Research Student of Mechanical Engineering of Agricultural Machinery, Urmia University, Urmia, Iran

[**] Department of the Mechanical Agricultural Machinery Engineering, Urmia University, Urmia, Iran

Email: hamid_majidi2007@yahoo.com

## Abstract

Apple is one the remarkable fresh fruit which contains a high degree of nutritious and medicinal values. Hand harvesting of apples by seasonal farm workers increases physical damages on the surface of this fruits, which causes a great loss in marketing quality. The main objective of this study is focused on designing a robust vision algorithm for robotic apple harvesters. The proposed algorithm is able to recognize and classify 4-classes of objects found in an orchard scene including: apples, leaves, trunk and branches, and sky into two apples and non-apples classes. 100 digital images of *Red Delicious apples* and 100 digital images of *Golden Delicious apples* were selected among 1000 captured images of apples from 18 apple's gardens in West Azerbaijan, Iran. An image processing algorithm is proposed for segmentation and extraction of the image's classes based on the color characteristics of mentioned classes. *Invariant-Momentums* were chosen as the extracted features from the segmented classes, e.g. apples. *Multilayer Feedforward Neural Networks, MFNNs,* were used as an artificial intelligent tool for recognition and classification of image's classes.

# 1. Introduction

Apple harvesting by seasonal fruit pickers is one the main causes of physical damages on the surface of apples. Presentations of physical damages like bruises on the fruits' surface decrease their marketing quality and endanger producers' place in the world marketing. In addition to the mentioned problems, matters related to workers' safety and fees from the one hand and decreasing the number of seasonal workers from agriculture part and joining to industry especially in the developed countries, European countries, has faced the producers with main difficulties. Latter is as a result of the economical advantages of industry in compared with agriculture. Moreover, fruit picking is a temporary operation. In the meantime, increasing the cost of food production in developed countries in compared with low-developed countries has caused the apples' producers are not to able to compete in world marketing cycle [1].

Therefore, considering the associated matters reveals the fact that automating the picking phase of fresh fruits like apples is the best solution to cope with aforementioned problems. Of course, it is worth to mention that automating of fruit picking operation is a complex and difficult process because robotic harvesters utilizes artificial intelligent systems in which computer vision put itself at the top of this technology. Therefore, designing a machine vision algorithm for recognition and classification of apples in an orchard scene is the challenging part of the robotic apple harvesters.

There have been many published reports on fruit recognition systems in recent years. Zhang [2] presented a machine vision algorithm in order to cucumber recognition for a robotic harvester. They applied a 3-layer B.P neural network to segment the cucumber's plant images. The $B^1$ and $S^2$ components were chose as the neural network inputs in their study. After successful training of the network all the cucumber's images were segmented. Finally, after texture analysis, they used third momentum as the extracted feature to identify the upper part of the fruit. The experimental results on 40 cucumber's plant images showed a 76% accuracy of recognition. Bulanon [3], proposed a machine vision algorithm for recognition of Fuji apples on the tree. They used luminance and color difference transformations to segment apples from the background. Whitaker [4], designed a machine vision system to identify green tomatoes, proposed algorithm was based on the shape information. Slaughter and Harrel [5], presented a vision algorithm to locate mature oranges based on color images. Their proposed method utilizes Hue and Saturation components of the HSI color model for analysis. Regunathon and Lee [6], implemented a machine vision system which had comprised from a color camera for imaging and an Ultrasonic sensor to measuring distance between camera and tree. The core of their algorithm for segmenting of citrus from background of acquired images was based on three classifiers: Bayesian, Fischer and watershed transformations. Grand D' Esnon [7], generalized a vision system for the MAGALI, a fruit picker robot, to identify apples in color images. At the core of their system, there was an analog signal processing system that was able to mark color points in a given color images. The proposed vision system was suffered from two major problems. First, it was not able to distinct sky between the leaves of tree. Second, it was necessary to use a dark background in the behind of tree during the imaging time, which made the imaging process more complex.

---
[1] Blue
[2] Saturation

The main aim of the presented paper is to design and implement a vision algorithm in order to identifying and classification of the objects of Red Delicious and Golden Delicious apples' gardens. The proposed algorithm is compatible with the physical characteristics of the Urmia gardens.

This paper is organized as follows: In Section 2, we describe material and methods applied in order to conduct this study, such as: some of the preprocessing operations, e.g. for size reduction and noise suppression and the proposed image segmentation method implemented in this study and the classification. In Section 3, we will regard results and findings obtained in this research and finally, section 4 is allocated to conclusion.

## 2. Materials and Methods

### 2.1. Data acquisition

Image capturing operation was done by a digital camera, Canon IXUS65 3CCD. Approximately, 500 digital images were captured from *Golden Delicious* apples and 500 from *Red Delicious* apples. We used a 64-Bit computer with a 2.5 GB memory in order to storing, analyzing and proposing vision algorithm and neural networks. MATLAB programming platform was used for implementing the proposed algorithm. The proposed algorithm was implemented for two favorable apples' varieties; *Red* and *Golden Delicious* apples. These both type are the two prominent varieties which have grown by farmers in West Azerbaijan, Iran. 18 standard apple gardens, with enough space between two rows and between trees on a row so that a hypothesis robotic harvester machine can move and work. Imaging sites located in *Barandoz* . Imaging was done during 5-days in September and 2-days in November 2008. In order to enhance the adaptability of proposed vision algorithm with the variation of the light intensity of the environment, images were captured between 8:30 a.m. and 6 p.m. Thus, the brightness of the captured-images was varying from dark-appearance to bright-appearance which were representative of the cloudy weather to sunny weather, respectively. Moreover, in order to simulating of the working condition of a hypothesis harvester images were captured from various perspectives and distances such as: far and close-distance or in the direction of up to down and vice versa. Fig. 1 shows some of these pictures. Finally, after evaluating 1000 images, we selected 100 images for *Red Delicious* and 100 images for *Golden Delicious apples* which were suitable for analyzing.

## 2.2 *Principle of designing a versatile vision algorithm for identifying apples*

We used color characteristics of image's classes; including: apples, leaves, sky and trunk-branches, in order to proposing a vision algorithm such that enable us to identify and segment apples from non-apple classes. *RGB Color space* was used for this purpose and all manipulating was done in this color space. This proposed algorithm decomposes all images into their constituting classes such as: apples, leaves, green objects, sky and trunk-branches. It is worth to mention that we define green objects as leaves and grass, therefore we classified leaves and grass as green class. In addition to image segmentation performed by this proposed vision algorithm, this vision algorithm assigns a constant gray-level value for each image's class. Therefore, this enabled us to control the fluctuating gray-level values which change abruptly in the dynamic range of 0 to 255. On the other words, the input of this algorithm is an RGB image with gray-level values between 0 to 255 for each *Red, Green* and *Blue* channels and output is a gray-level image with four constant intensity for each class, which considerably alleviates the difficulties of processing.

Briefly, the proposed vision algorithm consists of the 4 following steps:

1) Extraction of color characteristics of image's classes,

2) Preprocessing,

3) Main processing (Image segmentation plus forming of a 4 gray-level image),

4) Classification.

### *2.2.1. Color characteristic extraction of the 4-classes in images*

In this section, we manipulated the color characteristics of the Red Delicious apples with details and each-step images, but, since the method applied for extraction other class is in some similar to Red Delicious apples therefore, we would not mention them here. We only discuss non-Red-apples color characteristics in the section 5.

### *2.3.2. Color characteristics extraction of the Red Delicious*

Fig. 2 shows one of the 100 RGB color images of the *Red Delicious* apples processed in this study. After color sampling along the yellow-horizontal line from the surface of the *Red Delicious* apple shown in Fig. 2, corresponding color profile would be according to Fig. 3. As can be implied from RGB color profile of Fig.3 , the gray level values of the located pixels on the *Red component* of the RGB color space have the greatest gray level values with respect to located-pixels on the *Green* and *Blue* components. Approximately, color profiles of all *Red Delicious* apples in 100 color images have the similar locating order of *Red, Green* and *Blue* color components as shown in Fig. 3. It is worth to mention that *Blue* and *Green* components are located with a close distance with respect to each other. Fig. 4 shows the images of the *Red, Green* and *Blue* components of the Fig. 2 which proves the color characteristics of the *Red Delicious* apples. As can be seen from Fig. 4, *Red Delicious* apples have high gray level values, Fig. 4a. , in the image of the *Red Component* of the RGB Color space than *Green* and *Blue* components of the RGB color space, Fig. 4b and Fig. 4c. Therefore, it is natural that *Red Delicious* apples in the gray level image of the *Red Component* of the RGB color space appear brighter than two other components, *Green* and *Blue*. Color image profiles of *Golden Delicious apples,* leaves, and sky and trunk-branches class apples were computed as well. As mentioned in the earlier part of this section, we only mention their color characteristics in the discussion section**.**

## *2.4. Preprocessing*

In computer vision applications it is common to perform some preprocessing operations prior to main analysis. In this study we did some preprocessing algorithms such as: size reduction, noise suppression, data type transformations and histogram equalization.

The real size of the captured images were 2816 × 2112 pixels. Therefore, they required large memory, moreover, this large size of images led to the increase of the processing time and decrease of the performance. Accordingly, all images size was reduced to 1000 × 750 pixels.

All acquired-images from the *Red* and *Golden Delicious* apples have the data class of the uint8 in which the dynamic ranges of the gray-level values are in the range [ 0 , 255]. But, in weathering condition data class of images were changed into double data type, therefore, the accuracy of the gray-level values enhanced significantly.

In captured images especially in the cloudy weathering condition or images which captured in the evening, images had dark appearance; hence, most details were not obvious and clear. As could be seen from their histogram the gray-level values distribution of pixels were focused at the lower part of the histogram, which substantiate the dark appearance of the mentioned images. In order to rendering this situation , we applied histogram-equalization algorithm prior to processing of images captured in the dark time of the day or cloudy weather. According to the histogram of equalized images, gray-level values distribution of pixels was uniform, Therefore, the quality of the images enhanced appropriately and most details within images revealed.

In this study, we applied the Gaussian low pass filters to remove noises as unwanted data from the images. We passed each image through Gaussian low pass filter twice in order to noise suppression. Fig. 5a shows a binary image in which its gray-level image had not passed through filter and a binary image which its gray-level image have passed through a Gaussian low pass filter twice. As can be seen from Fig. 5b improvement is apparent.

## *2.5. Image Analysis*

After the color characteristics extraction of the images' classes, we segmented images into their constituent parts. As we mentioned in the earlier parts, our main aim is to segment and separate images classes and finally to form a gray-level image with 4 constant intensity values for each classes. In other words, instead of having an image with 256 gray-level values for each class, we constitute a gray-level image so that each class, e.g. apples class, has a constant gray-level value. To accomplish this task, the first step of the proposed vision algorithm is to transform RGB color images to their corresponding gray-level images and assign a constant pre-specified gray intensity for them. For example, Fig. 6a and Fig.6b shows an RGB color image of a *Red Delicious* apple tree and its corresponding gray-level image as a sample of the processed images in this study. From this point now, we will call Fig. 6b as "*host-image*" because after segmenting each class overlapping will done on this image.

### *2.5.1. Extraction of the Red and Golden Delicious apples*

In section 2.3.1 we computed the color profiles of the *Red Delicious* apples. As implied from the formed color profiles of *Red Delicious* apples, e.g. Fig. 3, the gray-level values of the located-pixels on the Red component of the RGB color space have the greatest values with respect to Blue and Green components of the RGB color space, respectively. Accordingly, there is no surprised in observing the image of the Red components of the RGB color space brighter than those images of the Blue and Green components, Fig. 4. Based on these information, we deduced that a *Color Difference Image* ,(CDI), which obtained from array subtraction of the Red components from the Green components resulted a gray-level image of the *Red Delicious* apples so that apples have the high gray-level values in that image. In mathematical denotation:

$$fDRG = fR - fG \qquad (1)$$

Where: *fDRG* is a gray-level image that has obtained from the subtraction of the Red component from the Green component, *fR* is the Red component and *fG* is the Green component of the RGB color space. Fig. 7a shows the *fDRG* image and Fig. 7b shows the color difference image obtained from the subtracting of the Red component from the Blue component. Thus, by comparing Figs. 7 a and Fig. 7b, and also by considering associated color profiles, it is obvious that Fig. 7a is a good image for apple segmentation. In addition, the histogram of the *fDRG* is a Bi-modal histogram; hence, *fDRG* images are as appropriate candidates for applying the automatic Otsu thresholding algorithm for the detection of *Red Delicious* apples. At the histograms of the *fDRG* images, lower mode belong to the non-apple class, represented by darker gray-level values and second mode belongs to the apples' class, represented by the brighter gray-level values in the *fDRG* images and these two modes are separated by a distinct valley as a threshold value for thresholding operation. After the thresholding from Fig. 7a by a threshold obtained from Otsu algorithm, a binary segmented image formed so that apples have the binary 1s values and represented by white color and non-apple classes have 0s and shown by black color as background, Fig. 8a.

After obtaining the segmented images of the *Red Delicious* apples, we overlapped this image, Fig. 8, with the "*Host-Image*", Fig. 6b, by the overlapping algorithm. At the overlapping phase, we assigned the constant gray-level value of 255 for the *Red Delicious* apples in the gray-level Host-Image that completely correspond to the segmented apples in the segmented binary image. Finally, after overlapping, the result image was according to Fig. 8b. As cab be seen from Fig. 8b all *Red Delicious* apples have the constant gray-level values of 255, and seen in the white color. Of course, it is worth to mention that, sky in Fig. 9 has gray-level value of the 255 as well. This problem will be address in the extraction of the sky class.

*Golden Delicious* apples, leaves and Blue sky segmented with a same method that *Red Delicious* apples extracted above. Therefore, because of the similarity of the method we avoid to explaining them in detail and only mention it concisely. In brief, for segmenting and isolating *Golden Delicious* apples based on the information of their color profiles, the best candidate image for applying Otsu thresholding algorithm in order to segmentation *Golden Delicious* apples were the subtracted image resulted from the subtracting of the Red component from the Blue component in the RGB color space. On the other hand, mathematical representation image for Golden Delicious apples segmentation is according to:

$$fDRB = fR - fB \qquad (2)$$

Where: *fDRB* is the image obtained from subtracting of the Red components from the Blue components, *fR* is the Red component and *fB* is e Blue component of the RGB color model. Similarly, in order to extract and segment the leaves class using color characteristics the best candidate image obtains from the subtraction of the Green component from the Blue component. Finally, for segmenting and extracting Blue sky class, sunny weathering condition, according to the color profiles of the blue sky, the best candidate image was a bi-modal histogram image that was obtained from the subtraction of the *Blue* component from the *Red* component in RGB color space. Mathematical denotation for representing candidates' images for segmenting of leaves and Blue sky classes are same as Eq. 1 and 2. Thresholding and overlapping algorithms for extracting of the *Golden Delicious*, leaves and Blue sky classes are completely similar to the methods applied for *Red Delicious* apples with the exception that the assigned constant gray-level values in the gray-level host-image in the overlapping phase is 180 for the leaves and 60 for the Blue sky class.

## 2.5.2. Extraction of the cloudy sky class

The color of the sky is in the direct relation with the meteorological condition. When the weathering condition is cloudy, sky's color will be seen in the white color at the acquired images and according to their color profiles, there is only Blue component in its color profiles. The first method that may come into mind at the first glance is that using Blue component of the RGB color space of cloudy sky and applying direct thresholding on it would give a good solution, Fig. 9a. As can be seen from Fig. 9a, applying this method revealed some parts of the *Red Delicious* apples and also branches with cloudy sky class. Therefore, although this method extracted cloudy sky on one hand, but on the other hand, it would lead us to misclassification problems. In our proposed method, we first subtracted Red component from the Green component and then computed the complement of this image. Finally, applying the automatic Otsu thresholding algorithm extracted the sky with high accuracy. It is worth to note that we changed the data type of the cloudy sky images from uint8, un-signed integer values in the range [0 255], to double data class before applying set algebra to this kinds of images.

*2.5.3. Extraction of the trunk and branches class*

As mentioned in above sections, color characteristics of the trunk and branches are approximately corresponds with the color characteristics of the *Red Delicious* and leaves classes. Accordingly, the applied method for extracting and segmenting of the apples, leaves and Blue sky classes would not lead to a correct solution. Therefore, like cloudy sky class, we used set theory to identifying this class. Up to this point, we have identified and extracted three classes; including: apples, leaves and sky from total four-classes. Thus, the complement of a gray-level image with three pre-specified class would be the complement class; trunk and branches class. We assigned the constant gray-level value of zero for cloudy sky in the gray-level host-image in the overlapping phase. Fig. 9b, shows a final output image with four-specified classes in which all apples have the constant gray-level values of 255, leaves with the gray-level values of the 180, sky with constant gray-level values of 60, trunk and branches with the gray-level values of the zero. This processed photo is the output of the image processing part of the proposed vision algorithm developed in this study.

*2.6. Feature Extraction*

Since inserting images as classifiers' inputs is difficult for processing by computers, therefore, we extracted some numerical descriptors from the processed images and presented them as classifier's inputs. In this study, we used *Artificial Neural Networks* as classifiers. In this study we selected invariant momentums, the extracted features, in order to describing the images' classes. The 2-D momentum of order (p + q) of a digital image, *f(x, y,)* is defined as below [8]:

$$m_{pq} = \sum\sum x^p y^q f(x,y) \qquad (3)$$

The invariant momentums defined according to Eq. 4 for digital images:

$$\eta_{pq} = \frac{\mu_{pq}}{\mu_{00}} = \frac{\sum\sum(x-\bar{x})^p(y-\bar{y})^q f(x,y)}{\mu_{00}^{\gamma}} \qquad (4)$$

Where:

$\eta_{pq}$, is the normalized central momentum of order (p + q),

$\mu_{pq}$, is the central momentum,

$\mu_{00}$, is the momentum of order 0, average of the image,

$\gamma = (p + q)/2 + 1$, $p + q = 2, 3, \ldots$ ,

$\bar{x} = \frac{m_{10}}{m_{00}}$, and $\bar{y} = \frac{m_{01}}{m_{00}}$.

After implementation of the invariants momentums in the MATLAB programming environment, we examined the various combination of the invariant momentums of orders 1st to 7th, and finally, we selected 3×1 feature vectors for describing images' classes. The sole criteria for choosing feature vectors were the increasing of the discriminatory measure between images' classes so that the classifiers can distinguish each class from others. Finally, we selected the following feature vectors for the classes of the 200 digital images.

*1) Apples' class:* defined by a 3×1 feature vector in which its first element is the invariant momentum of order 3, second element is the invariant momentum of order first and third element is the invariant momentum of order second, or in the vector representation:

$$\text{Apple Feature Vector} = \begin{bmatrix} \text{Momentum 3rd} \\ \text{Momentum 1st} \\ \text{Momentum 2nd} \end{bmatrix}$$

*2) Leaves class (Green object class):* The best descriptor for defining this class was a 3×1 feature vector in which the first, second and third elements are the invariant momentums of order first, fourth and third, or in the vector denotation:

$$\text{Leaves Feature Vector} \begin{bmatrix} \text{Momentum 1st} \\ \text{Momentum 4th} \\ \text{Momentum 3rd} \end{bmatrix}$$

*3) Sky class:* A 3×1 feature vector in which the first, second and third elements of it are the invariant momentums of order fifth, third and fourth, respectively.

$$\text{Sky Feature Vector} = \begin{bmatrix} \text{Mometum 5th} \\ \text{Momentum 3rd} \\ \text{Momentum 4th} \end{bmatrix}$$

*4) Trunk and Branches class:* since the gray-level values, *f(x, y)*, of this class is the zero at the output image of the image processing part of the vision algorithm, hence, all elements of the feature vector of this class is zero, or:

$$\text{Trunk and Branches Feature Vector} = \begin{bmatrix} 0 \\ 0 \\ 0 \end{bmatrix}$$

Feature extraction phase were done on the 4-classes of the all 200 digital images of *Red Delicious* and *Golden Delicious* apples.

## *2.7. Classification*

In order to recognition and classification of the images' classes into two apple and non-apple classes, we used multilayer feedforward neural networks trained with backpropagation learning rules. Thus, the intelligence part of the proposed vision algorithm would enable a robotic apple harvester to recognize and classify objects of an orchard scene either as apples or non-apples. The inputs of the neural networks were the extracted 3×1 feature vectors that were calculated at the section 2.6. Accordingly, for 200 images of the *Red Delicious* and *Golden Delicious* apples approximately, 800 3×1 feature vectors presented to the neural networks. We trained the neural networks so that it classified apples in the binary 1 class and non-apples in the binary zero class.

### *2.7.1. 2-layered Feedforward Neural Networks*

First, we used a 2-layered feedforward neural network for recognition and classification of the images' classes. The transfer functions of the neurons were sigmoid functions in each layer. At first, 20 neurons were put at the hidden layer. In order to training, 70 percent of inputs were used for training, 15 percent for validation and 15 percent for simulating of the networks after training phase. Neural's learning rule was the gradient back propagation learning rule with gradient descent method. But, unfortunately, the classification error of this network was about 26.6 percent which was not a satisfactorily result. Even, after increasing of the number of neurons in the hidden layer to the 40, 60, 80 and 100, the classification accuracy of the network did not increase from the 74 percent.

### *2.7.2. 3-layered Feedforward Neural Networks*

In order to increasing the recognition and classification accuracy of the neural networks in section 2.7.1, we formed the 3-layered feedforward neural networks with 2-hidden layers and 1-output layer. We trained 3-layered neural networks with two backpropagation learning rule; gradient descent and momentum learning rule. We designed thirteen 3-layered neural architectures based on the neurons on their hidden layers; Table 1. The main criteria for selecting the optimized architecture from Table.1 were based on the minimum classification error. In training, we presented 60 percent of the inputs for the training of the networks, 20 percent for the validation and 20 percent for the simulating of the trained network in all 13 structures, Table. 1. During the processing of data by the aforementioned networks, we recorded learning cycles and learning time need for convergence of the neural networks (Epochs), Regression coefficients for the training, validation and simulating phases and also for when all inputs presented to the neural networks simultaneously; we called this "whole data mode". Next, in order to increasing the learning speed, this time, we applied the backpropagation learning rule with momentum method on the high performance trained networks with the gradient descendant method.

## *3. Discussion and Results*

In this study, two varieties of Apples, *Golden Delicious* and *Red Delicious* apples have been studied using our designed and implemented intelligent machine vision algorithm. Different apple orchards were used to gather image samples. Multi-layer artificial neural networks were designed in order to recognition and classification of the images' classes. In this section, we consider and discuss our finding regarding vision algorithm design in this study.

### *3.1. Light Intensity*

During the pre-processing phase, we determined that the image qualities during the image acquisition phase were predominantly affected by light intensity due to the atmospheric conditions on the day of image acquisition in apple harvesting season. In a sunny day, the images were good and showed details. On cloudy days when the light intensity was low, the images lost details. For this purpose, we integrated a histogram equalization scheme in our pre-processing stage to improve the latter images. Essentially, the efficacy of any classification algorithm in this type of work depends heavily on the quality of image processing within the system. The harvest time of the Red Delicious apple is earlier in season than that of the Golden Delicious', we made sure that we take our image samples during similar good sunny days and time of the day between 8:30 a.m. to 6:00 p.m. to reduce light variation bias in our samples. Furthermore, we incorporated an adaptive ability for image enhancement with respect to the acquired image brightness and contrast. Applying these kinds of images increased the reliability of the proposed algorithm in confront with unpredicted situations which are common during harvesting season.

### *3.2. Results of the Color characteristics of 4-Classes of the Images*

With color sampling of the Red Delicious apples which are red colored, we concluded that the gray level intensity profile of the pixels on the color image's red component has the highest values saturated near the top, then comes that of the green's followed by blue's. In mathematical representation,

$$R>G>B. \quad (5)$$

As we saw in the profiles shown on Fig. 3, the green and blue profiles are quite close, even sometimes fall over each other. During image acquisition stage, it was concluded that the blue and green color components as opposed to that of the red's are highly sensitive to image lighting conditions. For example, in Fig. 10a, sunlight is shining over some of apples in the picture: As shown in Fig.10b, for Apple #1, the RGB blue component profile shoots up to near that of the red components. On the other hand, the less the light intensity, such as being in the shadow or on a cloudy day, the blue component profile of our RGB model drops to low values and distances itself from the red component's.

For the class of the Golden Delicious apple, the color yellow of this type of apple produces very high but close to red and green profiles while that of the blue's is distant from both. It was also observed that Golden Delicious apples around the month of September when they are still unripe, are greenish and their green component profile supersedes that of its red's and both are much higher than that of the blue's, or in mathematical notation:

$$G>R>B \quad (6)$$

And for Blue sky in sunny weathering condition:

$$B > R > G \qquad\qquad (7)$$

For blue component, the gray level intensity values are generally above 200. On cloudy days, the sky color appears white and the color components contains only blue component with gray-level values around 255. The main trunk of the Red Delicious and Golden Delicious apple trees are nearly grayish in color while the branches are reddish in color. This double jeopardy in color attributes of trunk and branches caused us not to apply these features in our recognition design.

### *3.3. Image Preprocessing Results*

Image preprocessing is a fundamental phase of any image recognition design. In this section, the results of preprocessing conducted experiments are briefly discussed.

To conduct the entire recognition operation in real-time or near real-time, it became necessary to reduce the size of images because processing time was found to be geometrically proportional to the image sample sizes. The images acquired by the digital camera Canon IXUS65 were of dimensions 2816x2112 pixels. The latter dimensionality was too large for our tasks in processing and computer memory requirements. For example, if the input images were to be 2816x2112, the processing time required for would be 412 seconds (about 7 minutes) which would be prohibitively slow for real-time automation robotic applications of apple harvesting. We wanted to have high speed and accuracy at the same time. If the images were reduced to 1000x750, the processing time would accordingly be reduced to 12.79 seconds. It was determined that the use of 8 bit integers to represent gray levels between 0-255 does not provide the accuracy needed for our recognition tasks in segmentation of the cloudy sky class. We increased our gray scale dynamic range from integer to real and using that we performed our computations in double precision in the cloudy weathering condition. With trial & error, it was determined that applying Gaussian filtering 3 times successively over all gray scale image classes but only once on sky class would be optimum for noise reduction. More filtering repetitions would start eliminating signal data along with noise.

### *3.4. Extraction Results and Discussion for Red Delicious Apples*

As shown by Figures 2, 3 and, the red component profile from color sampling of Red Delicious apple surface contains highest gray level intensity followed by those of the green's and blue's. Keeping in mind that in thresholding, a good line of attack is from the deepest histogram valley (or dip), as can be seen from their histograms, the deepest valleys in all three of these histograms occur between 200 and 256. Thus, in image binarization stage, if the threshold value is chosen below 200, the binary images will not be accurate due to loss of details above 200 gray level intensity. Because of the multiple modality nature of the gray level of the red, green and blue components of a Red Delicious apple image, simple thresholding will not render adequate separation of the apples from other classes. Therefore, we resort to a form of adaptive thresholding. By revisiting the color component profiles of Red Delicious apples, we concluded that if the red component gray scale image is subtracted from that of the green's, another gray scale image will be created in which only the Red Delicious apple lighter silhouettes will stand out while other classes within the image, because of smallness of their gray level values, will look faded out of the picture, appearing much darker than the apple class. Figure 3 shows the Red Delicious profiles. As can be verified from Fig.3, the gray scale image obtained from subtraction of red from green color components will have higher gray level values than that obtained from subtraction of those of red from the blue component'. The resultant two images from the latter subtractions have been shown in Fig.7a and 7b. Thus, it was decided that the best way to bring out Red Delicious apples is subtractions of the red from green component intensity values; that of red from blue brings out some parts of other classes; e.g. the class of leaves. As such, in the Bi-modal histogram of the Fig. 7b existed two peaks and a deep valley between them, therefore, predicting the existence of 2 objects within an image, one (the first modal on the left) representing the dark object and two (the second modal) due to a light object (apple in this case). Bi-modal histograms are the best for automatic thresholding to extract the object within an image.

### *3.4.1. Potential Errors in classification of Images' objects*

One possible error occurs when the profile has been taken from the part of the images' classes on which sun is shining. This example, e. g. Red Delicious apples, has been marked by a cross (x sign) on the image below. In Fig. 10a, because of the sun rays, the blue profile will be the dominant profile (Fig. 10b). Fig.10c shows the binary image related to Fig.10b, where there is sun rays shining directly over the samples. For the sun ray condition in Fig. 10b depicted in RGB color space, using even the HSI color space which is insensitive to the variation of the environmental lighting conditions don't render satisfactory results; Fig. 10d. In Fig. 10d light reflectant parts are shown in blue color. Therefore, applying other color spaces does not render this problem anyway and the sun rayed portions often are erroneously classified as sky class. Occurring of this error in the extraction of the all images' classes is unavoidable. The only solution to this problem is to stop working for about 1.5 to 2 hours at the lunch time when sun is shining directly or use an umbriferous as an integrant part of the robotic harvesting machine. The same problem occurred in the proposed vision algorithm by Bulanon and Kataoka (2008) [9]. Fig. 11 shows this problem in our proposed; lower row, and Bulanon and Kataoka's vision algorithms; upper row. In Fig. 11, it is worth to mention that because of the structure difference between Our proposed vision algorithm and Bulanon and Kataoka's vision algorithm, apples in our proposed vision algorithm are depicted by binary 1s value, white color, and background is depicted by binary 0s values, black color, while in Bulanaon and Kataoka's algorithm these values are in converse. As can be seen from Fig. 11, sun-shinning parts are classified as the background. But, on the other hand, Bulanon and Kataoka's vision algorithm suffers from a big problem and it is only able to recognize big apples in the images, while our proposed algorithm obviates this limit.

The discussion and results concerning the extraction of the *Golden Delicious* apples, leaves , sky, trunk and branches classes are the same with *Red Delicious* apples discussed here. Thus, we do not mention them here again. Moreover, it is worth to mention that our proposed vision algorithm have another superiority over designed algorithms by Grand D' Enson, which was developed for the MAGALI robot harvester. His vision algorithm was not able to identify the sky class presented between leaves class and, hence, he used a black curtain for solving this problem which in turn made the adaptability of his proposed vision algorithm with the main difficulties. While, our proposed vision algorithm dose not suffer from this problem.

## *3.5. Feature Vector Selection*

In Real-Time or near Real-Time processing, direct inserting of the image itself into a classifier like artificial neural network, not only increases the processing time but also produces erroneous results. In this study, we examined several feature vectors for describing each images' class, but the best results obtained by the invariant momentums of size 3×1. However, in this study, some outlier data happened during feature selection. Considering the all numerical features of the 200 digital images, we got to this fact that at seven images, extracted features for sky class are positioned within the apples' class. The presence of the outlier data makes some misclassification errors in our classification stage. In the other words, sky class in seven image is classified as apple class. Referring to these images, in which outlier data phenomena happened, Fig. 12. We concluded that, this event happens in pictures that sky class presented in all over the images, e.g. images that were taken from the bottom toward up perspective. As implied from the name of momentum, it is a kind of moment in plane, therefore, at these kinds of images,  which the sky covers all over the image, the gravity center of the sky closes to the gravity center of the image, thus making the decreasing of the momentum values near the range of apples class invariant momentum. Of course, it is possible to prevent from the happening of this event; for example, by altering the imaging technique.

*3.6. Artificial Neural Networks*

The main purpose of the application of the neural networks as classifier in this study was the recognition and classification of the images' classes into apples and non-apples classes. In order to enhancing the network performance, we focused our main attempt on the increasing of the number of the presented data to the neural networks, because by this way, the networks experienced the large range of inputs which cause the networks be robust in performance. In this study, we applied 2-layer and 3-layer neural networks with different number of neurons at their hidden layers and we used two backpropagation learning rule: with gradient descent and momentum. 2-layer networks with different number of neurons at their hidden layers did not get a satisfactory result, on the other words; their recognition accuracy did not increase more than 74 percent, even when we increased the number of neurons to 100 at the hidden layer. Hence, we concluded that applying 2-layer networks would not be a appropriate choice in this research.

Application of the 3-layer neural networks was accompanied with desirable results. Our sole criteria in selecting the optimized neural networks from the networks' architectures in Table 1, were based on the minimum error function in company with high regression coefficient for training, validation, testing and whole data mode. Considering the recorded results of the applied neural networks based on the above criteria, the high performance network with backpropagation learning rule of gradient descent were 50-40-1 and 100-100-1 architectures. For the 50-40-1 structure, error function was 0.00182, network's accuracy was 99.82%, and regression coefficients for training, validation, testing and whole data mode were 99.85%, 99.96%, 99.99% and 99.91%. Similarly, for the 100-100-1 architecture, error function was 0.00681, network's accuracy was 99.32% and regression coefficients for the training, validation, testing and whole data mode were: 99.25%, 99.97%, 99.97% and 99.45%, respectively. In order to enhancing the speed of the neural networks, decreasing of the learning Epochs with the stipulation of the retaining of networks' accuracy, we developed the gradient descent to momentum learning rule. Since, learning with momentum rule was the generalized and developed version of the gradient descent method; therefore, we varied the momentum coefficients from 0.1 to 0.9 for the two optimized networks trained by gradient descent: 50-40-1 and 100-100-1. Results indicated that for the 50-40-1 structure with the momentum learning rule the highest speed reached at 23 seconds and with the coefficient momentum 0.3. However, at the other hand, its regression coefficient for the testing phase was zero for all momentum coefficients. Evaluation of the obtained results of training for 100-100-1 structure with momentum learning rule indicated that this network reached its highest speed with the high accuracy at the momentum 0.8, in which the classification accuracy was 98.98 percent and regression coefficients for the training, validation, testing and whole data mode were: 99.25, 99.69, 99.94 and 99.38 percent, respectively. In addition this network converged at 2049 Epochs and in 40 seconds. Finally, regarding the results of the applications of the neural networks in this study, we concluded that the robust neural network in this study, such that it be applicable in a hypothetical robotic harvester, for classification and recognition of apples in an orchard scene is the 100-100-1 structure trained with backpropagation learning rule with momentum coefficient 0.8. Comparing the accuracy of the applied classifiers by the other researchers such as: Zhang et al: 76%; Whitaker et al: 68%: Buemi: 90%; Regunathon and Lee: 64.2, 73.2 and 68.8% for Bayesian, Neural Networks and Fischer classifiers, respectively; with the accuracy of the proposed neural networks in this study revealed the fact high classification accuracy of our proposed classifier indebted to the training of the network with the more inputs and the appropriate structure of the proposed neural networks. For example, applied neural networks at Yang et al. vision system for recognition and classification of the weeds from young corn trained with the data of 80 images while we used the data of 200 images for training of our proposed neural network applied in this research.

## *4. Conclusion*

We proposed a vision algorithm for recognition and classification of *Red Delicious* and *Golden Delicious* apples applicable in a hypothetical robotic apple harvester so that it can be able to recognize and classify apples in an orchard scene. Although, we processed our algorithm in off-line mode, but, we attempted to simulate a Real-Time working conditions, so that a hypothetical automated harvester would face with a high probability. We concluded that the best characteristics for analyzing images in order to extract images' classes is the color characteristics of the images' objects in the RGB color space. In order to segmenting images' classes, we applied Otsu automatic thresholding algorithm. Invariant momentums were used as the best descriptors for defining images' classes. 3-layer neural network with the backpropagation learning rule and with momentum 0.8 was applied in the proposed vision algorithm in order to classify images classes to apples and non-apples classes. It is worth to mention that stopping working at noon's when sun shines directly to the scene or using an umbriferous will prevent the misclassification errors, and finally, by altering imaging technique so that they were not taken from bottom to up direction, so that the sky was not presented all over the image, will obviated the occurring of the outlier data phenomena.

**Table. 1. Different Artificial Neural structures used in this study**

| Architecture | Number of Neurons (first hidden layer) | Number of Neurons (second hidden layer) | Number of Neurons (output layer) |
|---|---|---|---|
| Structure 1 | 4 | 10 | 1 |
| Structure 2 | 10 | 20 | 1 |
| Structure 3 | 40 | 40 | 1 |
| Structure 4 | 40 | 50 | 1 |
| Structure 5 | 50 | 40 | 1 |
| Structure 6 | 40 | 60 | 1 |
| Structure 7 | 60 | 40 | 1 |
| Structure 8 | 80 | 60 | 1 |
| Structure 9 | 60 | 80 | 1 |
| Structure 10 | 80 | 80 | 1 |
| Structure 11 | 80 | 100 | 1 |
| Structure 12 | 100 | 80 | 1 |
| Structure 13 | 100 | 100 | 1 |

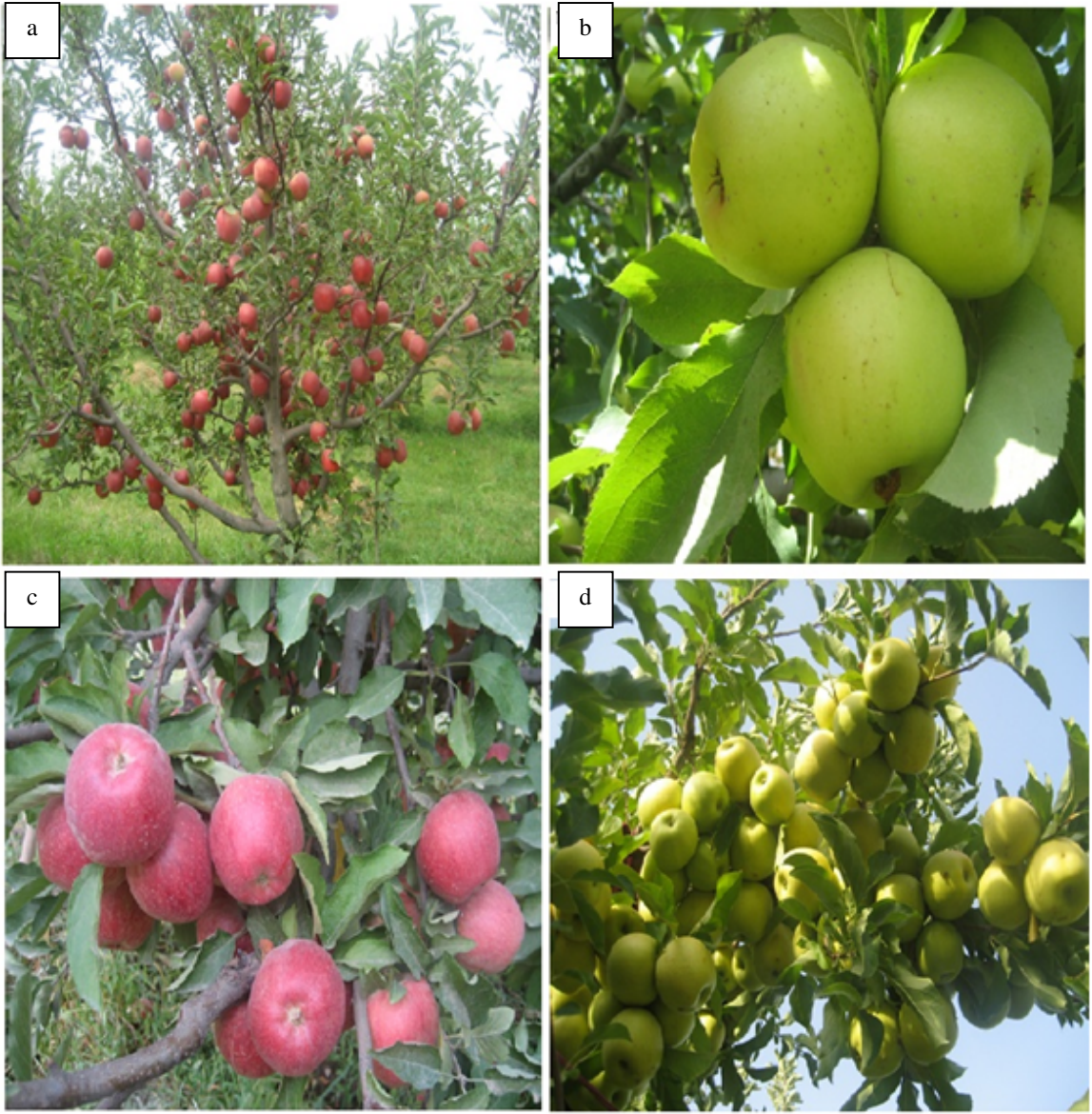

**Fig. 1. a) far-distance image, b) close-distance image, c) up-to-down image and d) down-to-up image**

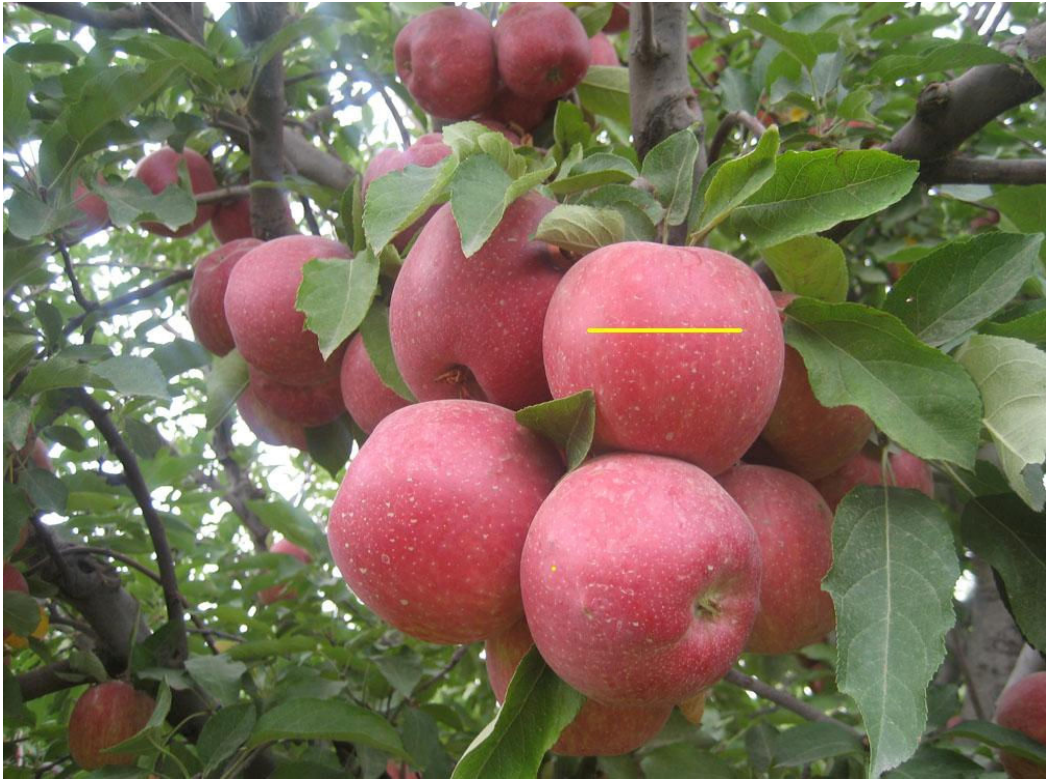

**Fig. 2. RGB color image of a Red Delicious apple tree with yellow-marked line for extraction of the color characteristics.**

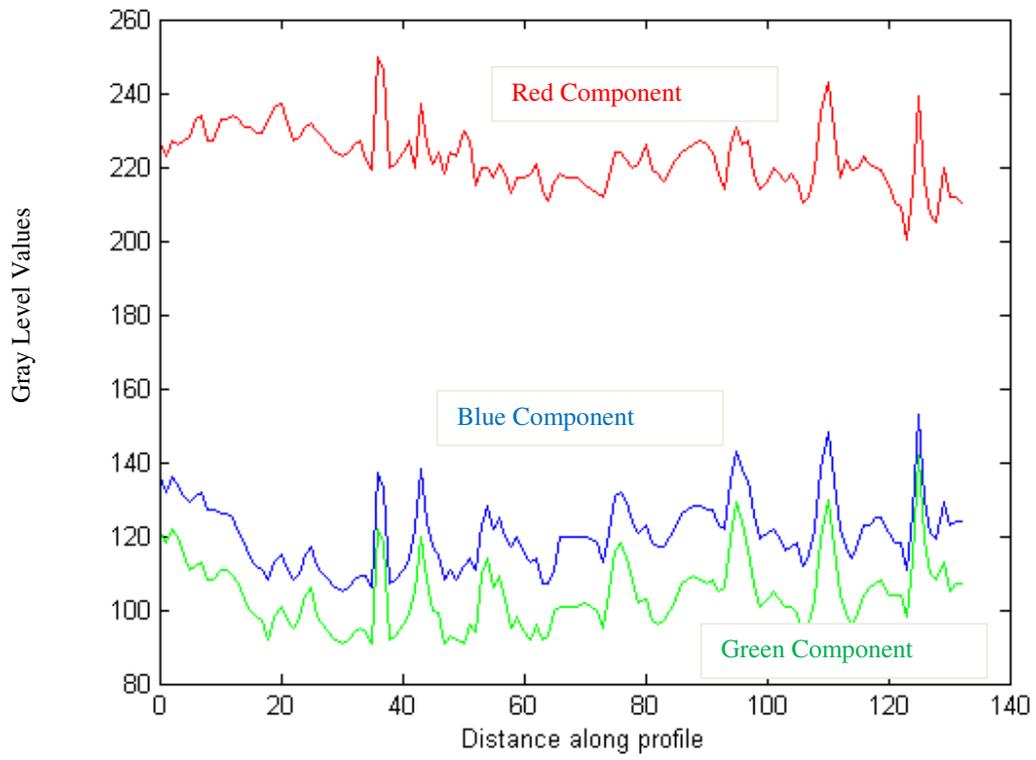

**Fig. 3. Color profile of Red Delicious apple; shown in Fig. 2.**

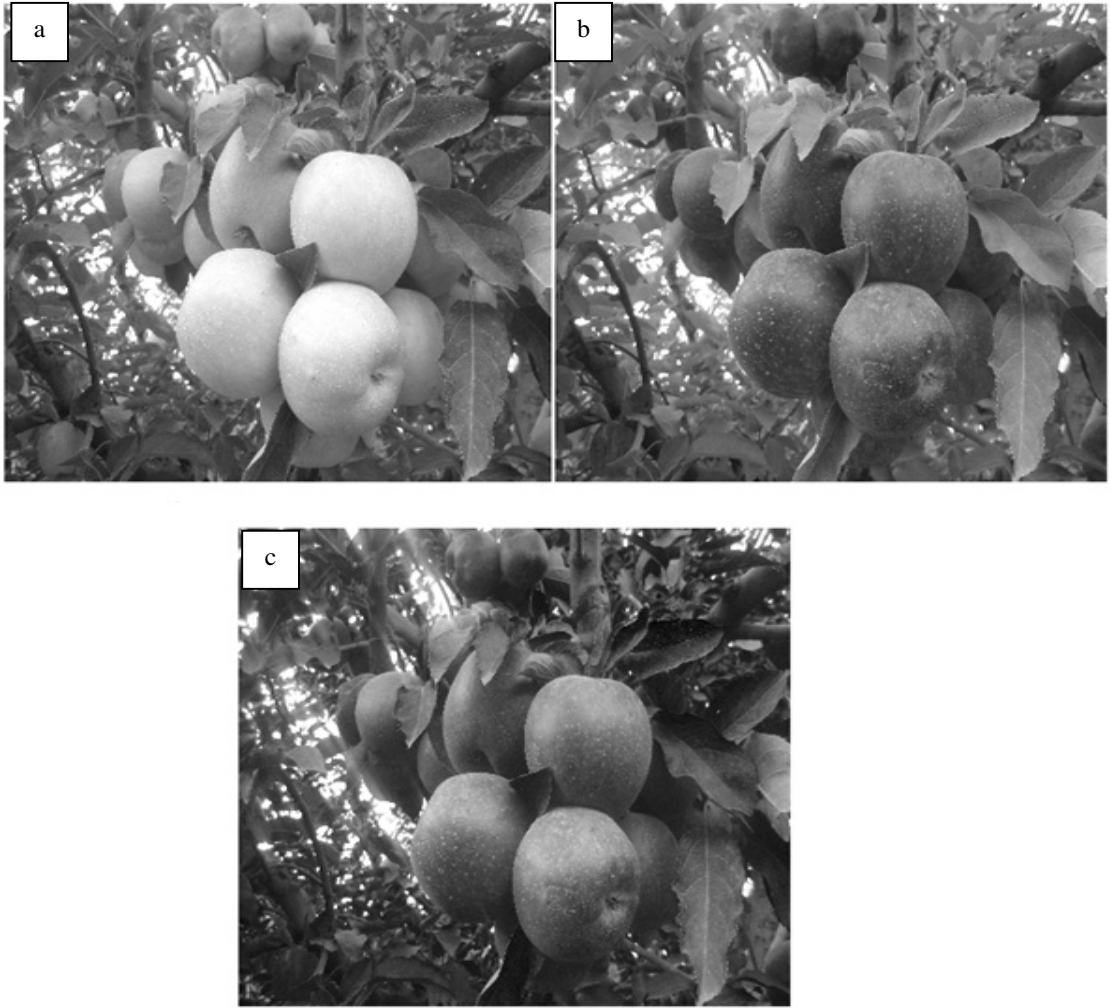

**Fig. 4. a) image of the *Red Component*, b) image of the *Green component* and c) image of *Blue component* of the *Red Delicious* apple tree shown in Fig. 2.**

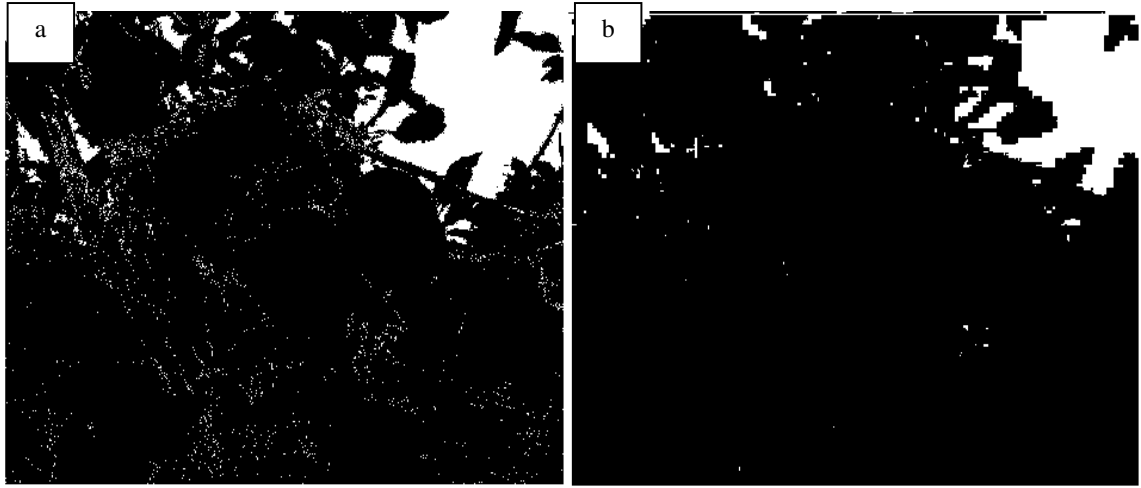

**Fig.5. a) unfiltered image, b) filtered image with Gaussian low pass filter**

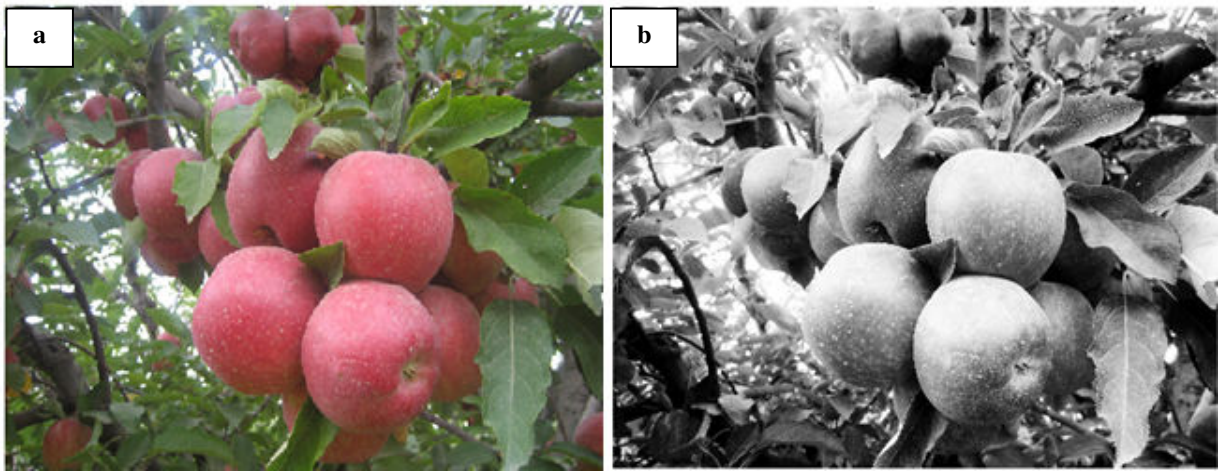

**Fig. 6. a) An RGB-Color image of a *Red Delicious* apple tree, b) gray-level image of the Fig. 6.a**

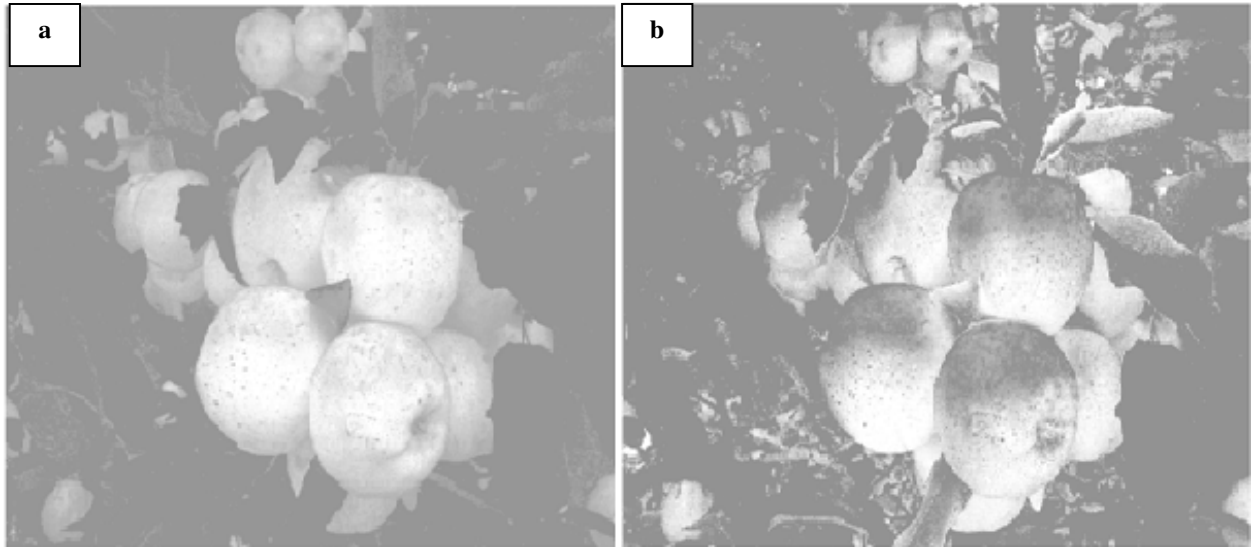

**Fig. 7. a) Color difference image obtained from the subtraction of the Red components from Green component and b) Color difference image obtained from the subtraction of the Red component from the Blue component**

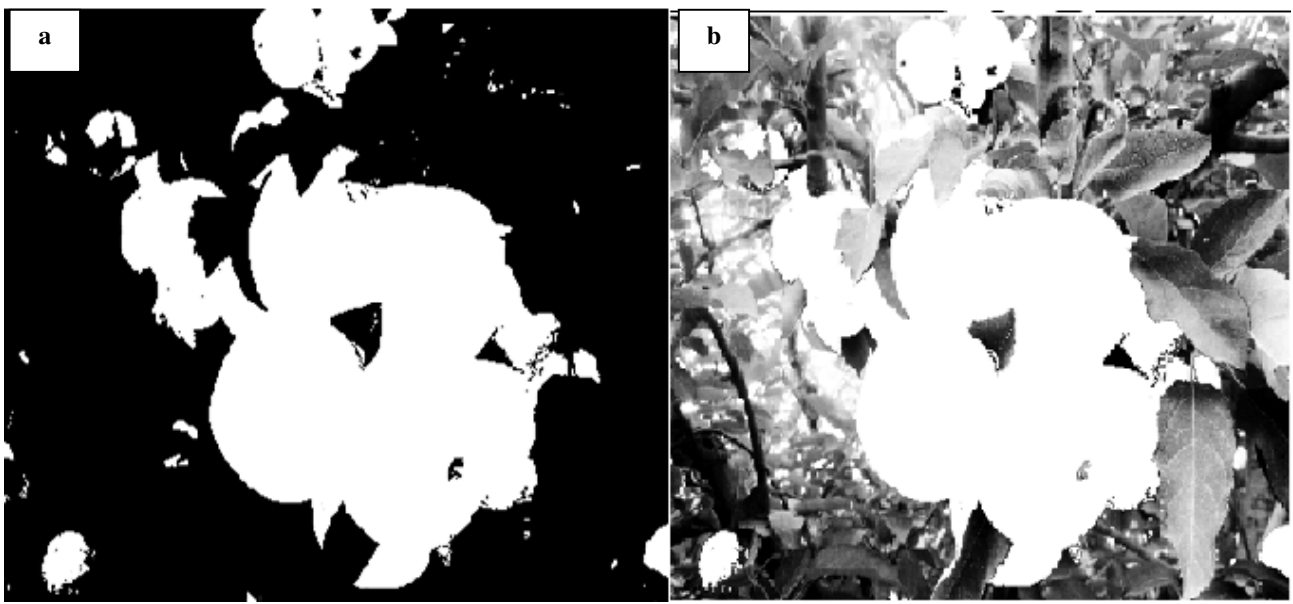

**Fig. 8. a) Segmented image of the *Red Delicious* apples and b) Resulted gray-level image obtained from overlapping algorithm**

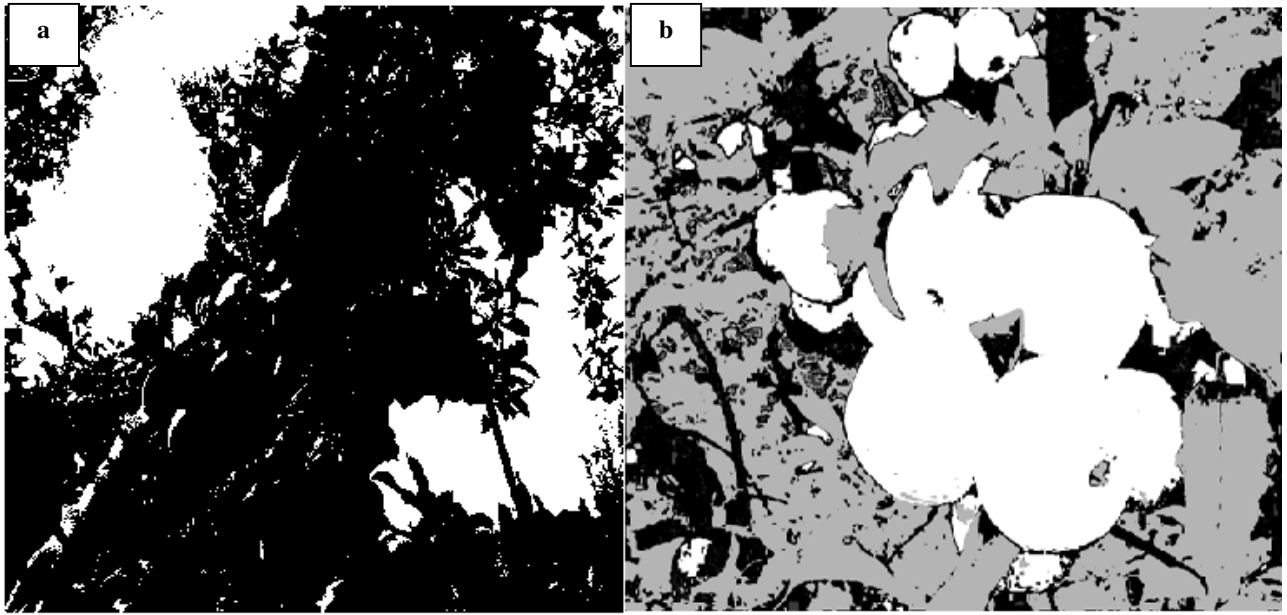

**Fig. 9. a) Segmented image obtained by applying direct Otsu thresholding on Blue components of Cloudy sky and**

**b) Fianl gray-level image with four-specified classes**

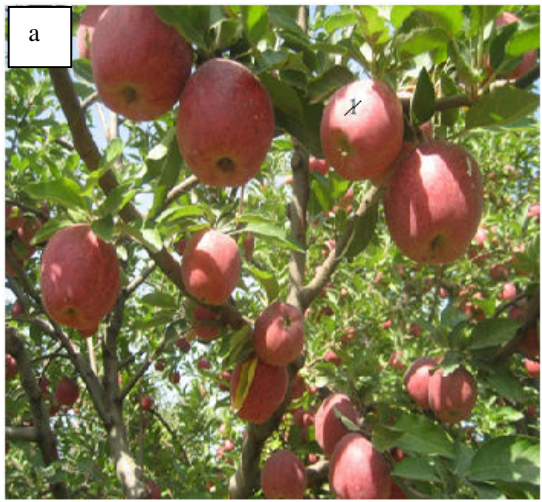
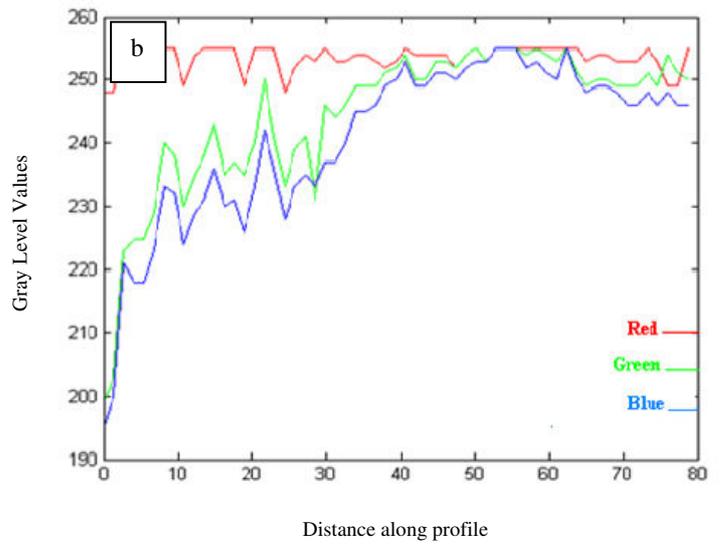
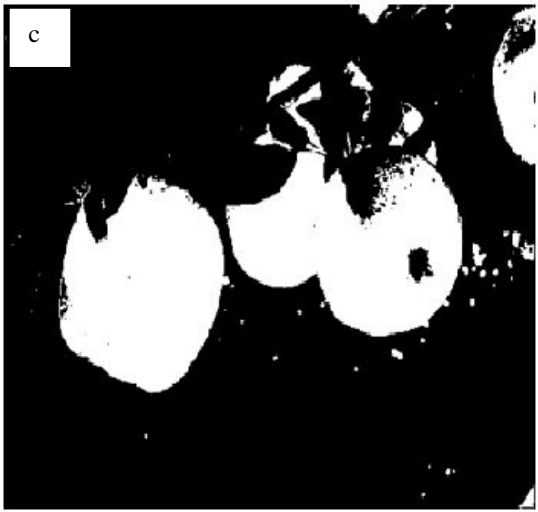
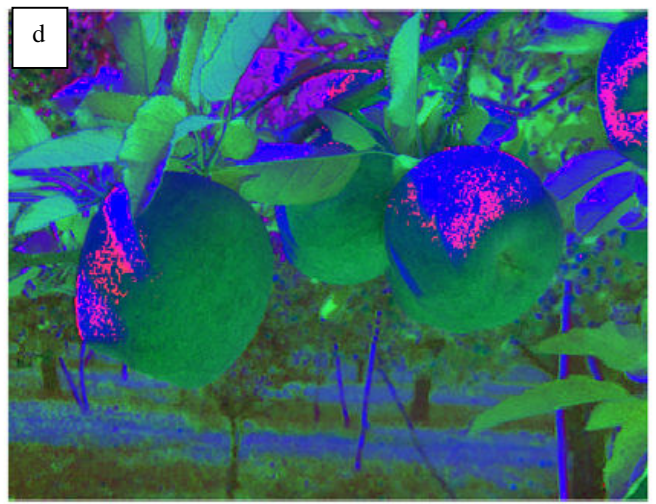

**Fig. 10. a) Red Delicious Apples under Sunlight, b) Profiles of Apple #1 in Fig. 10a, c) . Binary Image for Fig. 10a and d) HIS Representation of Fig. 10. a**

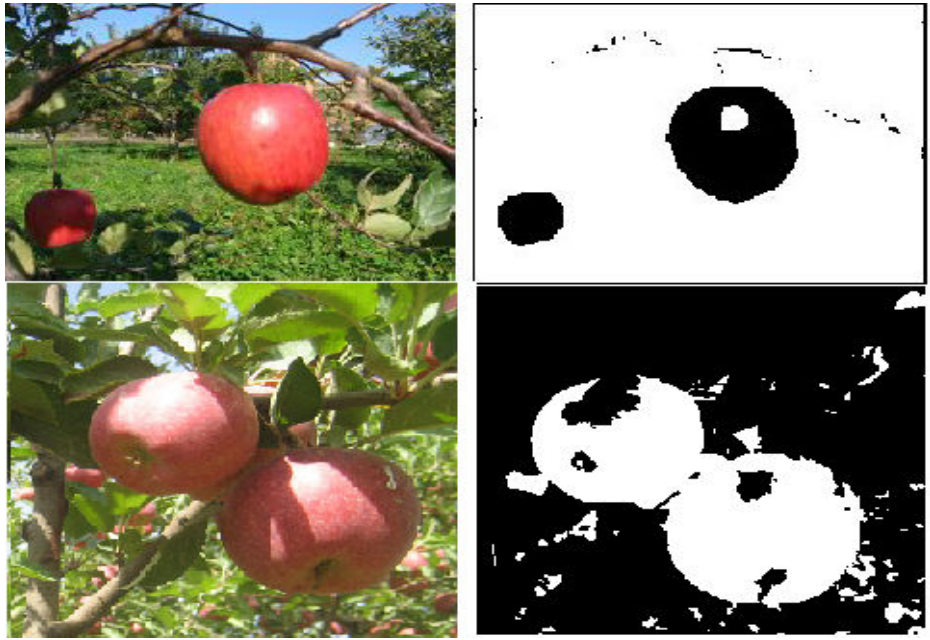

**Fig. 11.** *Upper row***: misclassification error in the Bulanon and Kataoka vision algorithm and** *lower row***: same problem, error, in our proposed vision algorithm.**

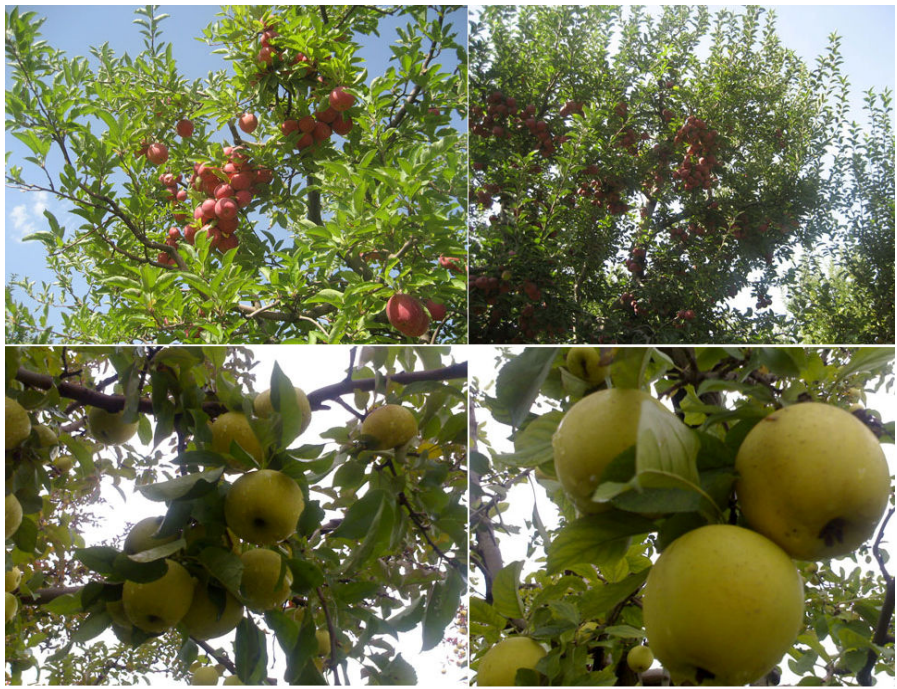

**Fig. 12. Some of the images in which outlier data phenomena happened**